# Benchmark Granularity and Model Robustness for Image-Text Retrieval

**Mariya Hendriksen*** 
University of Amsterdam
Amsterdam, The Netherlands
m.hendriksen@uva.nl

**Shuo Zhang**
Bloomberg
London, UK
szhang611@bloomberg.net

**Ridho Reinanda**
Bloomberg
London, UK
rreinanda@bloomberg.net

**Mohamed Yahya**
Bloomberg
London, UK
myahya6@bloomberg.net

**Edgar Meij**
Bloomberg
London, UK
emeij@bloomberg.net

**Maarten de Rijke**
University of Amsterdam
Amsterdam, The Netherlands
m.derijke@uva.nl

## Abstract
Image-Text Retrieval (ITR) systems are central to multimodal information access, with Vision-Language Models (VLMs) showing strong performance on standard benchmarks. However, these benchmarks predominantly rely on coarse-grained annotations, limiting their ability to reveal how models perform under real-world conditions, where query granularity varies. Motivated by this gap, we examine how dataset granularity and query perturbations affect retrieval performance and robustness across four architecturally diverse VLMs (ALIGN, AltCLIP, CLIP, and GroupViT). Using both standard benchmarks (MS-COCO, Flickr30k) and their fine-grained variants, we show that richer captions consistently enhance retrieval, especially in text-to-image tasks, where we observe an average improvement of 16.23%, compared to 6.44% in image-to-text. To assess robustness, we introduce a taxonomy of perturbations and conduct extensive experiments, revealing that while perturbations typically degrade performance, they can also unexpectedly improve retrieval, exposing nuanced model behaviors. Notably, word order emerges as a critical factor – contradicting prior assumptions of model insensitivity to it. Our results highlight variation in model robustness and a dataset-dependent relationship between caption granularity and perturbation sensitivity and emphasize the necessity of evaluating models on datasets of varying granularity.

---

*Work done while interning at Bloomberg AI.

## 1 Introduction

Image-text retrieval (ITR) is a bidirectional retrieval task that involves retrieving the top-$k$ relevant images given a textual query—or vice versa—based on cross-modal semantic alignment [4, 24]. This capability plays an important role in multimodal information access, enabling more expressive search and interaction paradigms [9]. Fueled by advances in large-scale pretraining, vision-language models (VLMs) have achieved state-of-the-art (SOTA) performance on standard ITR benchmarks [15, 38, 59, 73].

*The granularity gap.* The ITR landscape has largely relied on established benchmarks such as MS-COCO [13, 43] and Flickr30k [77] to evaluate model performance. While these datasets have been instrumental in driving progress, they predominantly employ *coarse-grained* captions, i.e., general descriptions that may overlook finer details important for assessing retrieval accuracy [12, 20, 34]. For example, a caption might describe "a person walking a dog" without specifying important distinguishing features like the breed of dog, the setting, or the person's attire. This level of abstraction may mask significant differences in models' ability to capture fine-grained semantic relationships.

Recent work has attempted to address this limitation through fine-grained dataset augmentations (MS-COCO-FG and Flickr30k-FG), which provide more detailed and descriptive captions [12]. However, to the best of our knowledge, the impact of this increased granularity on model performance and robustness remains poorly understood. Critically, we lack systematic studies examining how caption detail affects retrieval quality across different model architectures and evaluation scenarios.

*The robustness challenge.* Beyond granularity, *robustness* remains a critical yet underexplored aspect of the ITR evaluation. Real-world applications of ITR encounter noisy, ambiguous, and perturbed inputs – ranging from minor textual modifications to variations in image content [8]. Prior studies emphasize the necessity of models that generalize beyond clean, well-annotated benchmarks and maintain robustness against adversarial shifts and data perturbations [44–46, 48, 54, 55]. Furthermore, existing ITR evaluation metrics often rely on binary matches between images and texts, ignoring real-world scenarios where there may be partial semantic overlaps [50, 68, 82]. A more comprehensive assessment framework is needed to evaluate models' sensitivity to textual perturbations and their ability to maintain retrieval performance under real-world conditions.

*Research goals.* Motivated by this gap, in this reproducibility study, we aim to validate and extend previous findings on the role of dataset granularity and robustness in ITR evaluation. To ensure comprehensive evaluation, we select four architecturally diverse pre-trained models which demonstrate state-of-the-art (SOTA) performance on ITR tasks and evaluate them in a zero-shot setting. Following the ACM terminology [3], we focus on the *replicability* (different team, different experimental setup) of previously reported results. Following Voorhees [66], we evaluate the relative performance of VLMs on different datasets. We explore the impact of concept granularity in the context of robustness and extend beyond traditional binary matching to measure semantic alignment in cross-modal retrieval tasks.

In our experiments, we are motivated by the following research questions: (RQ1) How using more detailed (fine-grained) image descriptions affects retrieval performance compared to general (coarse-grained) descriptions across selected models (ALIGN, AltCLIP, CLIP, and GroupViT)? (RQ2) How is the performance of the selected state-of-the-art VLMs (ALIGN, AltCLIP, CLIP, and GroupViT) on the coarse-grained vs. fine-grained datasets impacted by perturbations? To address these questions, we conducted over 200 experiments testing 13 different perturbations across four selected models and four datasets. The experiments are grouped into two principal sets, each aimed at addressing a specific research question.

*Main contributions.* Our main contributions are: (1) We conduct one of the first reproducibility studies examining both dataset granularity and model robustness in ITR, by replicating experiments from [12] and extending them to analyze their generalizability. (2) We develop a comprehensive evaluation framework that systematically assesses VLM robustness to 13 different perturbations across both coarse-grained and fine-grained datasets, revealing unexpected cases where perturbations can improve retrieval performance. (3) We introduce a novel evaluation suite that bridges the gap between concept granularity and model robustness in ITR tasks. This suite provides: (i) zero-shot evaluation across multiple architecturally diverse models that excel at ITR, (ii) systematic analysis of perturbation impacts on both coarse and fine-grained datasets, (iii) cross-modal evaluation metrics that capture nuanced performance differences.

## 2 Preliminaries

*Notation.* We adopt the notation used in [7]. Let $\mathcal{D}$ be a dataset of $N$ image-text tuples: $\mathcal{D} = \{(\mathbf{x}_I^i, \{\mathbf{x}_{C_j}^i\}_{j=1}^k)\}_{i=1}^N$. Each tuple $i \in N$ consists of a single image $\mathbf{x}_I^i$ and $k$ corresponding texts (captions) $\mathbf{x}_{C_j}^i$, where $1 \leq j \leq k$. All texts are considered relevant to the image $\mathbf{x}_I^i$. We derive sets of queries $Q$ and candidates $C$ from the dataset $\mathcal{D}$. Let $Q_\mathcal{T}$ represent the set of text queries, where $Q_\mathcal{T} \subseteq Q$. Let $Q_I$ represent the set of image queries, where $Q_I \subseteq Q$. Similarly, $C_\mathcal{T} \subseteq C$ and $Q_I \subseteq Q$ represent the sets of text and image candidates respectively. Let $q \in Q$ and $c \in C$ represent a query and a candidate item respectively.

A query $q$ may originate from either the text modality $q \in Q_\mathcal{T}$ or the image modality $q \in Q_I$, while a candidate $c$ may similarly originate from either the text modality $c \in C_\mathcal{T}$ or the image modality $c \in C_I$. Let $E_{\theta_1} : Q \to \mathbb{R}^d$ be the encoder function mapping textual queries $q \in Q_\mathcal{T}$ to $d$-dimensional vectors: $\mathbf{q} = E_{\theta_1}(q)$. Similarly, we write $E_{\theta_2} : C \to \mathbb{R}^d$ for the encoder function mapping image queries $c \in C_I$ to $d$-dimensional vectors: $\mathbf{c} = E_{\theta_2}(c)$.

Let $f_{rel} : Q \times C \to \mathbb{R}$ be a relevance function that computes the relevance of a query-candidate pair. We write $f_S : Q \times C \to \mathbb{R}$ for a scoring function that takes a query and a candidate, maps them into $d$-dimensional space, normalizes the vectors so that they lie on $d$-dimensional hypersphere and computes their similarity. Finally, $f_{sim} : \mathbb{R}^d \times \mathbb{R}^d \to \mathbb{R}$ denotes a similarity function that computes a similarity score between the two $d$-dimensional vectors. We assume that all vectors lie on the surface of a $d$-dimensional hypersphere. Formally, this implies that $\|q\| = \|c\| = 1$ where $\|\cdot\|$ denotes the Euclidean norm.

*Task.* We focus on the task of *cross-modal retrieval*, which involves matching queries in one modality (e.g., text or image) to candidates in a different modality.

The retrieval process can occur in two ways: (i) *text-to-image retrieval* (t2i): given a textual query $q \in Q_\mathcal{T}$ and a set of candidate images $C_I$, rank the images by their relevance to $q$; (ii) *image-to-text retrieval* (i2t): given an image query $q \in Q_I$ and a set of text candidates $C_\mathcal{T}$, rank the texts by their relevance to $q$. For both tasks, dedicated encoders are used to map images and texts into a shared $d$-dimensional representation space. Once encoded, we compute the similarity between the query and candidate in this shared space to derive relevance scores.

Performance is typically evaluated bidirectionally using Recall@k (R@k), where $k = \{1, 5, 10\}$, and the sum of recall (rsum).

## 3 Concept Granularity in Image-Text Retrieval Datasets

We start our study by examining the concept granularity in image-text retrieval (ITR) datasets, focusing on features that influence the specificity and richness of textual descriptions.

### 3.1 Selected Features

To analyze concept granularity in ITR datasets, we examine linguistic features at both the noun phrase (NP) level and the caption level. These features help quantify the level of detail and specificity in image descriptions.

*3.1.1 NP-level granularity.* This section discusses linguistic features contributing to NP granularity in captions.

*Modifiers of the noun.* Adjectives and complement phrases (CPs) provide details about objects in images [56, 80]. By quantifying these modifiers, we assess the detail and granularity associated with objects [42]. Specifically, we count the number of adjectives and CPs per identified noun in captions.

*Semantics: Concept depth.* Concept depth reflects the semantic understanding captured within individual concepts in captions, indicating a deeper comprehension of the depicted scene [75]. Datasets with deeper conceptual information offer more detailed descriptions of visual content [58]. We measure concept depth by calculating the minimum depth of the corresponding synsets, considering the maximum depth across all synsets associated with a word.

*Determiners: Articles, quantifiers.* The use of articles and quantifiers impacts the specificity of noun descriptions [30]. Analyzing their occurrences offers insights into the explicitness and precision of noun specifications. We quantify the occurrences of articles and quantifiers linked to identified nouns in captions.

3.1.2 *Caption-level granularity.* Next, we consider caption-level features.

*Caption length.* The character count of a caption indicates the amount of information conveyed [36]. Longer captions are likely to include more details, contributing to finer granularity. We measure the total word count for each caption.

*Number of words.* The total word count is indicative of caption richness [36]. A higher word count suggests a more elaborate description, signaling finer granularity. We count the total number of words in each caption.

*Semantic diversity of concepts per caption.* Concept diversity is essential for analyzing granularity within ITR datasets [30]. It reflects the range of ideas and semantic complexity captured in a caption. We compute the ratio of unique synonyms to the total word count in each caption.

## 3.2 Granularity Analysis

3.2.1 *Datasets.* We conduct our experiments on two widely used ITR datasets and their fine-grained variants:

*MS-COCO [43].* A large-scale dataset originally designed for object detection, segmentation, and captioning. It consists of 123,287 images and 616,435 captions, with each image annotated with five captions.

*Flickr30k [77].* An image caption corpus consisting of 31,783 images and 158,915 crowd-sourced captions, with each image annotated with five captions. This dataset is commonly used for image-text retrieval tasks.

*MS-COCO-FG [12].* A fine-grained extension of MS-COCO that enhances concept granularity for more detailed retrieval evaluations. It augments the original dataset with captions containing additional contextual details extracted from the associated images.

*Flickr30k-FG [12].* A fine-grained extension of Flickr30k, designed to improve retrieval performance on nuanced textual and visual details. Like MS-COCO-FG, it adds captions with additional contextual information.

For all datasets, we use the standard training, validation, and test splits as defined by Karpathy and Li [31].

3.2.2 *Results.* Table 1 presents the results of analyzing our datasets in terms of granularity. For Flickr30k vs. Flickr30k-FG, we observe a 21% increase in the number of concept phrases in the extended dataset. This indicates a richer description of scenes with additional details. The concept depth remains unchanged. While the fine-grained dataset offers more detailed descriptions, the semantic complexity of the concepts remains largely unchanged. Similarly, we note a 38% increase in the number of adjectives per caption in MS-COCO-FG over MS-COCO. This suggests a more descriptive and nuanced portrayal of visual content. The concept depth exhibits only a marginal increase, implying that the semantic understanding of concepts is slightly enhanced in the fine-grained version. Overall, the fine-grained datasets demonstrate higher scores across features than their standard counterparts. Thus, they offer more detailed and descriptive captions, amounting to improved granularity.

## 4 Evaluation Framework

To comprehensively evaluate VLMs robustness in the context of ITR, we present a novel evaluation framework. This framework includes a diverse set of perturbations and a cross-modal relevance metric to examine model performance.

### 4.1 Perturbations

To assess the robustness and performance of VLMs in ITR, we introduce a set of perturbations targeting word order sensitivity and resilience to noise in input. These perturbations are inspired by prior studies on the limitations of large language models in handling word order [25, 53, 57, 78] and noisy input [18, 29, 64, 71, 83].

4.1.1 *Word Order Sensitivity.* To assess a model's sensitivity to word order, we designed a series of perturbations to test how rearranging sentence elements impacts its ability to perform ITR. We focus on three levels: adjectives and nouns, trigrams, and complete captions. We assume that breaking word order will degrade the model's retrieval performance, as sentence structure is crucial for accurate cross-modal alignment. The perturbations are grouped into three categories based on the level of operations:

*Nouns and Adjectives.* We test the models ability to handle changes in the arrangement of descriptive elements by shuffling the order of nouns and adjective (*shuffle nouns and adjectives*); additionally, we examine model's ability to preserve essential details while other sentence elements are rearranged (*shuffle all words but nouns and adjectives*).

*Trigrams.* We evaluate the model's response to localized word order changes by randomly shuffling the word order within each trigram (*shuffle within trigrams*); besides, we assesses the model's ability to perform ITR when faced with trigram reshuffling (*shuffle trigrams*).

*Complete Caption.* We test the model's sensitivity to word order on a caption level by randomly reshuffling all words in a caption (*shuffle all words*).

4.1.2 *Robustness to Noise in Input.* To evaluate the robustness of VLMs to noise in input, we introduce several perturbations that simulate common real-world scenarios. These perturbations are designed to test the model's ability to handle distractions, lexical variations, and typos.

Table 1: Coarse-grained vs. fine-grained ITR datasets at the levels of noun phrases and captions. Section 3.1 defines the quantities counted for each of the features.

| Level | Aspect | Features | MS-COCO | MS-COCO-FG | Flickr30k | Flickr30k-FG |
|---|---|---|---|---|---|---|
| NP | Modifiers of the Noun | Adjectives | 0.76 | 1.05 | 1.14 | 1.3 |
| | | Complement Phrases | 1.56 | 1.99 | 1.81 | 2.19 |
| | Determiners | Articles | 2.14 | 2.34 | 2.27 | 2.55 |
| | | Quantifiers | 0.12 | 0.13 | 0.26 | 0.27 |
| | Semantics | Concept depth | 7.89 | 7.91 | 7.97 | 7.97 |
| Caption | Number of Characters | Caption length | 52.39 | 56.38 | 63.61 | 68.29 |
| | Number of Words | Number of words in a caption | 10.59 | 11.48 | 12.34 | 13.67 |
| | Semantics | Diversity of concepts per caption | 9.14 | 10.04 | 9.86 | 10.68 |

*Distractions.* Distraction-based perturbations aim to evaluate the model's robustness to irrelevant elements within captions. These perturbations focus on statements that are always true and do not add meaningful content to the caption, helping to understand how well the model can filter out relevant information when performing ITR [64].

*Lexical variations.* This type of perturbation aims to assess the model's adaptability and robustness to changes in language [18, 29]. We focus on replacing $k$ synonyms and nouns in a given caption with their lexical variations.

*Typos.* Typos are common in real-world ITR scenarios, and evaluating a model's response to such errors is important for ensuring its practical usability [61, 71, 83]. Typo perturbations aim to assess the model's resilience to typographical errors. We implement several perturbations of this type that simulate keyboard character transposition, mimic a character omission typo, simulate insertion typo, and emulate key proximity typo.

### 4.2 Evaluation Metric

Our goal is to evaluate not only explicit matches but also the overall relevance between queries and candidates, even when explicit labels are unavailable. To achieve this, we define a metric based on both *perfect match* cases and *cross-modal relevance*.

We operate in a setup when, given a query $q$, and a ranked list of top-$k$ retrieved results $K = [c^1, \ldots, c^k]$, we want to obtain a list of the relevance scores $[rel^1, \ldots, rel^k]$ where $rel^i$ denotes the relevance for the $i$-th retrieved candidate.

*4.2.1 Perfect match.* When explicit matching labels are available, we assign a relevance score of 1 to perfect matches. This applies to both text-to-image and image-to-text retrieval:

(i) *Text-to-Image Retrieval (t2i)*: The retrieved image $c \in C_I$ is considered a perfect match if it is the ground-truth image for query $q \in Q_T$:

$$f_{rel}(q,c) = 1 \text{ if } \exists i \in \mathbb{N} \text{ such that } q \in \{\mathbf{x}_{C_j}^i\}_{j=1}^k \wedge c = \mathbf{x}_I^i.$$

(ii) *Image-to-Text Retrieval (i2t)*: The retrieved caption $c \in C_T$ is considered a perfect match if it is the ground-truth caption for query $q \in Q_I$:

$$f_{rel}(q,c) = 1 \text{ if } \exists i \in \mathbb{N} \text{ such that } q = \mathbf{x}_I^i \wedge c \in \{\mathbf{x}_{C_j}^i\}_{j=1}^k.$$

*4.2.2 Cross-modal relevance.* When explicit labels are unavailable (i.e., no perfect matches exist), the relevance score is computed based on the similarity between the encoded query and candidate vectors. This approach allows us to measure how well the model aligns cross-modal pairs (text and images) in the shared representation space. The scoring function $f_S$ is defined as:

$$f_S(q, c, E_{\theta_1}, E_{\theta_2}) = \begin{cases} f_{sim}(E_{\theta_1}(q), E_{\theta_2}(c)) & \text{if } q \in Q_T \text{ and } c \in C_I \\ f_{sim}(E_{\theta_2}(q), E_{\theta_1}(c)) & \text{if } q \in Q_I \text{ and } c \in C_T. \end{cases}$$

We use cosine similarity as the similarity function: $f_{sim}(\mathbf{v}_1, \mathbf{v}_2) = \frac{\mathbf{v}_1 \cdot \mathbf{v}_2}{\|\mathbf{v}_1\| \|\mathbf{v}_2\|}$.

*4.2.3 Overall metric.* To evaluate model performance across ranked results, we measure relevance while considering the rank position of the results:

$$DCG_{CM}^p = \sum_{i=1}^p \frac{\text{rel}^i}{\log_2(i+1)},$$

where $p$ represents the rank position up to which the score is computed.

## 5 Experiments

In this section, we describe the models selected for evaluation, the design of our experiments, and the results obtained. Our experiments aim to assess the impact of concept granularity on VLMs performance in ITR tasks and analyze their robustness to textual perturbations.

### 5.1 Models

For our experiments, we select four VLM that excel in ITR tasks. All selected models use dual-encoder architectures trained via contrastive learning, but each embodies distinct methodological approaches.

The models we consider are: (1) **ALIGN** [28] builds upon the principles established by CLIP (see below), using an expansive dataset of over one billion noisy image alt-text pairs. By using uncurated data, ALIGN achieves robust performance across large-scale visual tasks, distinguishing itself from models reliant on meticulously curated datasets and facilitating a more realistic assessment of robustness in less controlled environments. (2) **AltCLIP** [15] is a multilingual adaptation of CLIP (see below), enhancing its capabilities through the integration of a pre-trained multilingual text encoder, XLM-R, and a two-stage training schema that combines teacher learning and contrastive learning. This adaptation allows AltCLIP to achieve state-of-the-art performance on various vision-language tasks, demonstrating the effectiveness of simple modifications to CLIP's architecture for extending its capabilities in multilingual contexts. (3) **CLIP** [59] serves as a foundational model in VL research. It is contrastively pre-trained on a dataset of 400 million image-text pairs collected from the internet. CLIP's capacity for zero-shot transfer across a wide range of downstream computer vision tasks has established it as a benchmark in the field. Its efficient zero-shot performance provides a robust baseline. (4) **GroupViT** [73] features a hierarchical approach that focuses on grouping semantic regions within images without the need for pixel-level annotations. The model dynamically aligns image regions with their corresponding textual descriptions, emphasizing visual scene understanding by progressively grouping image regions into larger segments, which contrasts with the global image-level representations used by the other selected models.

## 5.2 Experimental Design

To answer the research questions introduced in Section 1, we conduct over two hundred experiments testing thirteen different perturbations across four selected models and four datasets. The experiments are grouped into two sets, each aimed at addressing a specific research question.

*5.2.1 Set 1: Coarse vs Fine-Grained Datasets Evaluation across Selected Models (RQ1).* In these experiment, we evaluate the impact of concept granularity in both textual descriptions and dataset composition on VLMs performance in the ITR task. We validate our evaluation framework by comparing our results to those reported in a previous study [12]. This study is relevant because it critiques current ITR benchmarks and proposes enhanced evaluations for fine-grained cross-modal semantic matching. Moreover, Chen et al. [12] introduced augmented benchmarks (MS-COCO-FG and Flickr30K-FG) that we incorporate into our experiments. We run the ITR task on both standard image-caption datasets (MS-COCO and Flickr30k) and their more fine-grained counterparts (MS-COCO-FG and Flickr30K-FG). The models are evaluated on image-to-text (i2t) and text-to-image (t2i) tasks, and we report the recall at 1 for both. This experiment allows us to assess how refining textual descriptions and increasing dataset granularity impact model performance.

*5.2.2 Set 2: Model Robustness and Perturbation Sensitivity (RQ2).* In these experiments, we the robustness to perturbations of state-of-the-art VLMs (ALIGN, AltCLIP, CLIP, and GroupViT) on the coarse-grained vs. fine-grained datasets. We apply 13 perturbations across the four selected datasets (MS-COCO vs. MS-COCO-FG, and Flickr30k vs. Flickr30K-FG). The perturbations are designed to test the models' sensitivity to changes in word order and robustness to noisy input. We analyze the performance drop of the models after each perturbation and measure their sensitivity to word order, lexical variations, and typos.

## 5.3 Results

*5.3.1 Set 1: Coarse vs. Fine-Grained Datasets Evaluation across Selected Models (RQ1).* To address RQ1, we evaluate models R@1 performance for both i2t and t2i retrieval and compare the results between the original datasets (MS-COCO, Flickr30k) and their fine-grained versions (MS-COCO-FG, Flickr30k-FG).

Table 2 highlights that refining the captions improves performance in most cases. Across datasets, we observe significant improvements in R@1 scores. The highest performance gain is a 29.11% improvement in CLIP for t2i retrieval on the Flickr30k dataset. On average, scores increase by 12.63% on MS-COCO and 10.05% on Flickr30k. Specifically, MS-COCO exhibits an 8.14% increase for i2t retrieval and a 17.11% increase for t2i, while Flickr30k shows a 4.75% rise in i2t scores and a 15.35% rise for t2i.

However, there are exceptions, particularly in the CLIP MS-COCO t2i and GroupViT MS-COCO i2t tasks, where refined captions do not improve the scores. Despite these few exceptions, the overall results demonstrate that refining textual descriptions enhances retrieval performance, with the greatest benefits observed in t2i retrieval, which saw an average 16.23% improvement compared to a 6.44% increase in i2t retrieval.

Therefore, we answer RQ1 as follows: fine-grained captioning consistently improves retrieval performance across models and datasets in zero-shot scenarios, with greater benefits observed in t2i retrieval than in i2t retrieval. The performance difference demonstrates that comprehensive model evaluation should include both granularity levels. Testing only on coarse-grained captions may underestimate a model's true retrieval capabilities, while testing only on fine-grained captions might overstate its real-world performance where detailed descriptions are not always available. This multi-granularity evaluation approach provides a more complete understanding of model robustness and capabilities across different levels of descriptive detail.

*5.3.2 Set 2: Model Robustness and Perturbation Sensitivity (RQ2).* To address RQ2, we assess the robustness of four VLMs (ALIGN, AltCLIP, CLIP, GroupViT) to various perturbations across MS-COCO, Flickr30k, and their refined counterparts. We apply the proposed perturbations to contrast how well models handle changes in word order, lexical variations, and typos, in the coarse-grained vs. fine-grained settings.

The results are shown in Table 2. The results indicate consistent drops across most perturbation-dataset pairs. The most notable decrease is caused by the *shuffle all words* perturbation, where randomly shuffling all words in captions leads to the largest score drops, underscoring the models' reliance on correct word order for accurate retrieval. In contrast, the *lexical variation* perturbation has the smallest effect, indicating a greater model resilience to synonym substitution. Interestingly, while most perturbations negatively affect performance, in some cases, refined datasets exhibit better robustness. For example, on MS-COCO-FG, models show smaller relative performance drops when compared to MS-COCO. This trend is less consistent for Flickr30k-FG, which shows smaller performance drops than Flickr30k for only 5 of the 13 perturbations. This discrepancy may be due to the inherently more detailed nature of Flickr30k captions, making additional granularity less beneficial than in MS-COCO, which has coarser captions (see Table 1 for details). Besides, while perturbations generally decrease performance across all models, we discovered several surprising cases where certain perturbations resulted in R@1 scores being improved. We present randomly sampled examples of these cases in the Appendix B. Table 4 illustrates two scenarios: one where perturbations increase R@1 and another where they decrease R@1. The left side of the table shows an example where a perturbation (changing "couple" to "coupel") led to an increase in R@1, as evidenced by the top-3 retrieved images. Conversely, the right side of the table demonstrates a case where a perturbation (changing "motorcycles" to "omtorcycles") resulted in a decrease in R@1, as seen in the corresponding top-3 images. Overall, our findings highlight the sensitivity of VLMs to perturbations, with word order being particularly critical. Interestingly, this contradicts prior work on this topic where authors demonstrate that reshuffling word order does not affect ITR performance [78].

Table 2: Model performance on the i2t and t2i tasks. "DCG" is short for "$DCG_{CM}$."

| Model | i2t | | | | t2i | | | | rsum | |
|---|---|---|---|---|---|---|---|---|---|---|
| | R@1 | R@5 | R@10 | DCG | R@1 | R@5 | R@10 | DCG | i2t | t2i |
| MS-COCO | | | | | | | | | | |
| ALIGN | 42.22 | 54.42 | 60.48 | 2.45 | 22.93 | 42.15 | 51.01 | 1.60 | 157.12 | 116.09 |
| AltCLIP | 40.95 | 53.44 | 58.64 | 2.43 | 22.47 | 41.85 | 50.90 | 1.61 | 153.03 | 115.22 |
| CLIP | 33.66 | 45.29 | 50.08 | 2.32 | 16.15 | 33.11 | 42.06 | 1.66 | 129.03 | 91.32 |
| GroupViT | 24.88 | 34.38 | 35.72 | 1.97 | 8.29 | 18.90 | 25.59 | 1.41 | 94.98 | 52.78 |
| MS-COCO-FG | | | | | | | | | | |
| ALIGN | 44.59 | 56.55 | 64.20 | 2.50 | 25.60 | 45.64 | 54.65 | 1.61 | 165.34 | 125.89 |
| AltCLIP | 43.97 | 57.23 | 61.83 | 2.51 | 25.45 | 45.86 | 54.75 | 1.63 | 163.03 | 126.06 |
| CLIP | 38.16 | 50.38 | 55.20 | 2.43 | 16.15 | 33.11 | 42.01 | 1.66 | 143.74 | 91.27 |
| GroupViT | 24.88 | 34.38 | 35.72 | 1.97 | 9.58 | 21.38 | 28.68 | 1.42 | 94.98 | 59.64 |
| Flickr30k | | | | | | | | | | |
| ALIGN | 70.52 | 83.58 | 88.90 | 3.03 | 35.56 | 58.78 | 67.64 | 1.70 | 243.00 | 161.98 |
| AltCLIP | 67.98 | 82.46 | 86.40 | 2.99 | 33.06 | 56.42 | 65.74 | 1.69 | 236.84 | 155.22 |
| CLIP | 58.06 | 72.54 | 79.30 | 2.85 | 19.30 | 39.74 | 49.22 | 1.70 | 209.90 | 108.26 |
| GroupViT | 35.34 | 49.24 | 50.80 | 2.20 | 8.36 | 19.26 | 26.02 | 1.38 | 135.38 | 53.64 |
| Flickr30k-FG | | | | | | | | | | |
| ALIGN | 75.28 | 87.38 | 90.80 | 3.10 | 39.80 | 64.76 | 73.44 | 1.73 | 253.46 | 178.00 |
| AltCLIP | 71.66 | 85.96 | 87.40 | 3.05 | 37.10 | 61.02 | 70.60 | 1.72 | 245.02 | 168.72 |
| CLIP | 63.70 | 77.72 | 82.60 | 2.95 | 24.92 | 46.00 | 55.60 | 1.73 | 224.02 | 126.52 |
| GroupViT | 38.50 | 53.88 | 52.30 | 2.26 | 8.92 | 20.98 | 28.54 | 1.38 | 144.68 | 58.44 |

Therefore, we answer RQ2 by stating that VLMs demonstrate varying degrees of sensitivity to different perturbations in zero-shot settings, with word order being the most critical factor affecting retrieval performance. More importantly, our findings emphasize the necessity of comprehensive perturbation testing in model evaluation, as these perturbations closely mirror real-world usage scenarios where input text is often imperfect. The interaction between caption granularity and perturbation robustness provides additional insights - models generally show better resilience to perturbations when operating on fine-grained descriptions, particularly in MS-COCO, suggesting that richer visual details in text may help maintain retrieval performance even under noisy conditions. This interplay between description granularity and perturbation robustness should be considered when developing and deploying VLMs in practical applications.

Overall, results underline the importance of comprehensive evaluation protocols for VLMs that go beyond standard benchmarks. The significant variations in model behavior across different granularity levels and perturbation types reveal capabilities and limitations that would remain hidden under simpler evaluation approaches. These findings suggest that robust assessment frameworks should account for both the quality of textual descriptions and the imperfect nature of real-world inputs to provide meaningful insights into model performance in practical applications.

## 6 Related Work

*Cross-modal retrieval.* Cross-Modal Retrieval (CMR) methods construct a multimodal representation space where concepts from different modalities are mapped and compared using distance metrics such as cosine or Euclidean distance. Early approaches relied on canonical correlation analysis [21, 32], followed by dual encoder architectures integrating recurrent and convolutional components trained with hinge loss [19, 69]. Later advancements introduced hard-negative mining [17] and attention mechanisms like dual attention and stacked cross-attention [35, 52]. Recent transformer-based methods leverage dual encoders trained on large-scale datasets. ALBEF [38] aligns unimodal representations before fusion, while CLIP [59] directly predicts image-text pairs. Models such as FILIP [76] and SLIP [51] enhance multimodal interaction and supervision techniques. AltCLIP [15] integrates multilingual text encoders, whereas GroupViT [73] incorporates a grouping mechanism in vision transformers to improve visual segment understanding. Unlike prior work in this domain, in our work, we conduct a comparative evaluation of multiple transformer-based dual encoder models on the image-text retrieval (ITR) task, analyzing their performance across different retrieval settings.

Table 3: Rsum after applying perturbation.

| Perturbation | MS-COCO | MS-COCO-FG | Flickr-30k | Flickr-30k-FG |
|---|---|---|---|---|
| **ALIGN** | | | | |
| No perturbation | 116.09 | 125.89 | 161.98 | 168.72 |
| Shuffle N&A | 100.00 | 109.58 | 139.33 | 145.39 |
| Shuffle all words | 85.78 | 97.58 | 120.39 | 130.77 |
| Shuffle all but N&A | 98.03 | 116.59 | 133.67 | 154.19 |
| Shuffle within trigrams | 101.70 | 116.12 | 144.65 | 154.16 |
| Shuffle trigrams | 104.23 | 117.86 | 145.06 | 156.83 |
| Distraction | 112.17 | 124.91 | 156.20 | 163.51 |
| Lexical variation | 108.88 | 119.46 | 157.79 | 161.61 |
| Typos | 103.07 | 115.25 | 152.83 | 152.01 |
| **AltCLIP** | | | | |
| No perturbation | 115.22 | 126.06 | 155.22 | 178.00 |
| Shuffle N&A | 96.84 | 107.54 | 133.63 | 154.82 |
| Shuffle all words | 88.41 | 98.91 | 121.62 | 132.39 |
| Shuffle all but N&A | 100.08 | 113.69 | 135.68 | 159.44 |
| Shuffle within trigrams | 101.60 | 113.66 | 138.82 | 160.87 |
| Shuffle trigrams | 103.81 | 115.35 | 143.14 | 163.60 |
| Distraction | 110.20 | 120.63 | 157.08 | 173.07 |
| Lexical variation | 107.46 | 118.20 | 148.64 | 174.12 |
| Typos | 100.91 | 112.60 | 141.32 | 161.71 |
| **CLIP** | | | | |
| No perturbation | 91.32 | 91.27 | 108.26 | 126.52 |
| Shuffle N&A | 31.23 | 72.24 | 86.06 | 99.74 |
| Shuffle all words | 41.24 | 60.87 | 69.19 | 77.82 |
| Shuffle all but N&A | 28.93 | 75.40 | 82.52 | 99.31 |
| Shuffle within trigrams | 26.11 | 74.12 | 84.57 | 100.26 |
| Shuffle trigrams | 30.60 | 76.41 | 91.08 | 103.33 |
| Distraction | 84.05 | 89.93 | 105.75 | 121.10 |
| Lexical variation | 74.12 | 84.04 | 101.26 | 139.32 |
| Typos | 66.30 | 76.86 | 87.99 | 105.37 |
| **GroupViT** | | | | |
| No perturbation | 52.78 | 59.64 | 53.64 | 58.44 |
| Shuffle N&A | 43.62 | 49.00 | 46.82 | 49.87 |
| Shuffle all words | 41.94 | 46.82 | 47.83 | 46.89 |
| Shuffle all but N&A | 49.08 | 54.58 | 51.82 | 48.32 |
| Shuffle within trigrams | 48.18 | 54.52 | 51.72 | 54.36 |
| Shuffle trigrams | 48.56 | 53.98 | 52.84 | 47.52 |
| Distraction | 51.18 | 58.23 | 53.47 | 59.91 |
| Lexical variation | 48.61 | 53.71 | 49.78 | 53.89 |
| Typos | 43.11 | 49.81 | 47.65 | 50.44 |

*Transformer-based vision-language models.* Another research direction explores transformer-based encoders for ITR. ViLBERT [47] and LXMERT [63] employ two-stream architectures, while B2T2 [2], VisualBERT [40], Unicoder-VL [37], VL-BERT [62], and UNITER [14] adopt single-stream architectures. Oscar [41] enhances region features by incorporating object tags, and BEIT-3 [70] extends multiway transformers trained with cross-entropy loss. This work focuses on transformer-based dual encoder models due to their strong performance on vision-language (VL) tasks. Unlike prior work in this domain, in our work, we systematically compare the effectiveness of four SOTA transformer-based dual encoders and provide insights into their generalization across different datasets.

*Vision-language model evaluation.* The evaluation of vision-language models (VLMs) is critical for assessing their capabilities across diverse tasks and datasets. Standard benchmarks such as MS-COCO [13, 43] and Flickr30k [77] have been widely used for image captioning, visual question answering (VQA), and ITR. However, these datasets have limitations in concept granularity and diversity, prompting the introduction of more fine-grained benchmarks like MS-COCO-FG and Flickr30k-FG [12].

Additionally, specialized datasets cater to specific domains: CUB-200 [72] for fine-grained bird classification, ABO [16] for product listings, and Fashion200k [22] for fashion items. Large-scale datasets such as Conceptual Captions [60], XMarket [6], and Recipe1M [49] further enrich the evaluation landscape, providing diverse real-world scenarios for testing VLMs. Unlike prior work in this domain, in our work, we evaluate these models on a broader range of datasets to analyze their performance in both standard and fine-grained retrieval tasks.

*Robustness and generalization.* Evaluating the robustness and generalization of VLMs is crucial for their deployment in real-world applications. Recent studies examine VLMs under adversarial attacks [81], domain shifts, and input perturbations [78] to identify vulnerabilities and improve model resilience. Adversarial attacks have been extensively studied in the context of VQA [5, 10, 33, 39, 67, 79] and image captioning [1, 11, 74], highlighting the need for robust training and evaluation strategies. Unlike prior work in this domain, in this work, we assess the robustness of transformer-based dual encoders under varying retrieval conditions in zero-shot settings and examine their performance in ITR scenarios.

## 7 Conclusions

In this work, we address the brittleness of the evaluation pipeline in the ITR task, emphasizing two primary concerns: the coarseness of existing benchmarks and the limitations of current evaluation metrics. Through our analysis, we compare standard datasets, MS-COCO and Flickr30k, with their fine-grained counterparts, MS-COCO-FG and Flickr30k-FG. We propose an evaluation framework that encompasses a taxonomy of perturbations and a new evaluation metric designed to improve the robustness of ITR assessments. We selected four state-of-the-art VLMs – AltCLIP, ALIGN, CLIP, and GroupViT – for our experiments and evaluate their performance on the ITR task using the novel framework.

*Main findings.* Overall, our findings reveal two critical aspects of VLM evaluation. First, the substantial performance differences between coarse and fine-grained datasets (particularly in t2i retrieval) demonstrate that comprehensive model assessment requires testing across multiple granularity levels. Second, the varying model responses to perturbations, coupled with the dataset-dependent relationship between caption granularity and robustness, highlight the complexity of real-world deployment scenarios. Models generally perform better with fine-grained descriptions but show dataset-specific patterns in their resilience to perturbations, suggesting that evaluation protocols should consider both description detail and text noise to better reflect practical usage conditions.

These insights underscore the importance of multi-faceted evaluation approaches that combine different granularity levels and perturbation types to fully understand model capabilities and limitations.

*Limitations.* While our study provides valuable insights, it has certain limitations. First, our evaluation focuses on a specific set of perturbations and datasets, which may not fully encompass the range of real-world variations encountered in image-text retrieval. Additionally, while we selected leading models in the domain of ITR, evaluating a broader range of VLMs could yield a more comprehensive understanding of their performance across diverse datasets and evaluation frameworks. Expanding our evaluation to include models with varied architectures and training methodologies could provide deeper insights into their robustness and generalization.

*Future work.* Promising avenues for future work include extending the proposed framework by incorporating additional perturbations and datasets, as well as expanding the range of evaluated models. Another promising avenue includes exploring other facets of VLM performance on the ITR task, such as interpretability and domain adaptation, to further improve our understanding of their capabilities and limitations.

## Reproducibility Statement

To ensure reproducibility and facilitate further research, we release our code at https://github.com/bloomberg/evaluating-cmr-in-mm. For our software stack, we employ Matplotlib [27] and SciPy for plotting, NumPy [23] for data handling, PIL [65] for image processing, and spaCy [26] for text processing. Regarding computational resources, all experiments were conducted on NVIDIA A100 GPUs (40GB memory). Evaluation runs used 1–8 GPUs for durations between 2–12 hours, depending on configuration. The total compute usage amounts to approximately 600 GPU days for experiments, with an additional 987 GPU days allocated for development.

## Acknowledgments

This research was (partially) supported by Ahold Delhaize, through AIRLab Amsterdam, the Dutch Research Council (NWO), under project numbers 024.004.022, NWA.1389.20.183, and KICH3.LTP.20.-006, and the European Union's Horizon Europe program under grant agreement No 101070212.

All content represents the opinion of the authors, which is not necessarily shared or endorsed by their respective employers and/or sponsors.

## Appendix
## A  Core Components

*Dataset.* The `Dataset` class handles loading and processing image-text pairs, supporting train/val/test splits, JSON annotation loading, and dataset augmentation. It maintains mappings between captions and images, applies augmentations when enabled, and retrieves image-caption pairs via indexing.

*Models.* The project supports multiple encoder architectures for computing image and text embeddings, including CLIP, ALIGN, AltCLIP, and GroupViT. Each model follows a common interface for encoding queries and computing similarity scores. Each model utilizes dedicated processor and backbone to encode images and text into a shared embedding space. We provide support for batch processing, with optimizations such as precomputed embedding storage and incremental computation when needed.

*Retrieval.* The `Retriever` class is responsible for retrieving the most relevant documents given a query. It takes a query, encodes it using the specified model, and computes similarity scores between the query and a set of document embeddings. The retrieval process begins by truncating textual queries to match the model's maximum sequence length if necessary. The query is then encoded into an embedding tensor, which is compared against the stored document embeddings using semantic similarity. The top-k most relevant documents are selected based on their similarity scores. The class returns the ranked document names and their scores.

*Evaluation.* The `Evaluator` class assesses retrieval performance through bidirectional ITR task. The evaluation process begins with dataset loading and preprocessing, followed by embedding computation, either from scratch or using precomputed values. The retrieval process then ranks candidate results based on semantic similarity scores, and performance is measured using metrics such as RecallK and $\text{DCG}_{CM}$.

*Metrics.* The `Metrics` package offers functionality for computing RecallK and the $\text{DCG}_{CM}$ metric. The $\text{DCG}_{CM}$ metric evaluates ranking quality by incorporating graded relevance scores, assigning perfect relevance to exact matches and computing partial cross-modal relevance for non-exact matches using the configured relevance estimator. The `RelevanceEstimator` class is responsible for computing relevance scores between queries and documents using CLIP-based models. It supports multiple model architectures and computes similarity scores using cosine similarity.

*Perturbations.* The `Perturbation` class applies various types of perturbations to captions to evaluate the robustness of models. The `TyposPerturbation` class introduces common typographical errors into captions. The `SynonymBased` class generates perturbations to test the model's ability to handle semantic variations. The `DistractionBased` class introduces distracting elements into the captions, aiming to test the model's focus and robustness against irrelevant information. The `ARO` class applies various perturbations to the captions, aiming to test model's sensitivity to word order.

Table 4: Examples of perturbation effects on R@1 for image retrieval.

| Perturbation increases R@1 | | Perturbation decreases R@1 | |
|---|---|---|---|
| *Initial caption* | *Perturbed caption* | *Initial caption* | *Perturbed caption* |
| a red rose is sitting next to a couple of mugs | a red rose is sitting next to a **coupel** of mugs | Two men are at an intersection on motorcycles. | Two men are at an intersection on **omtorcycles**. |
| Top-3 images | | Top-3 images | |

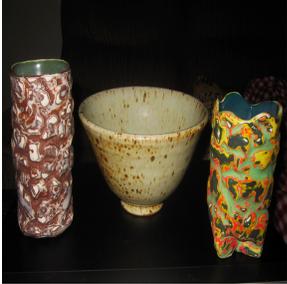 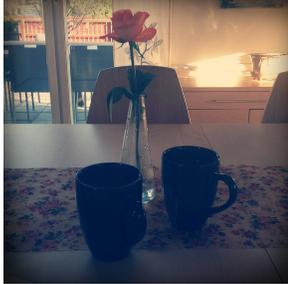 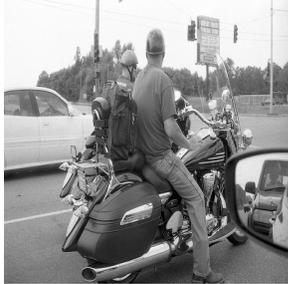 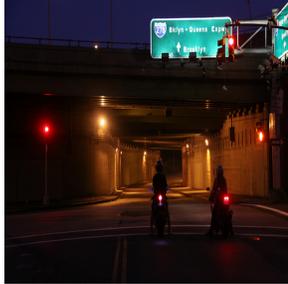

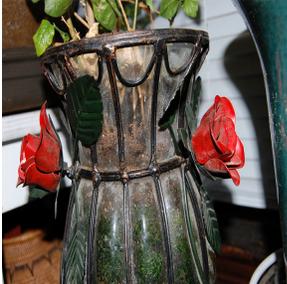  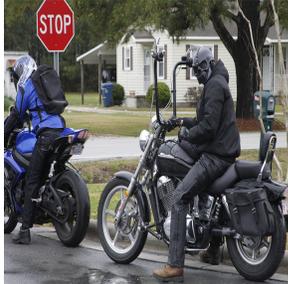 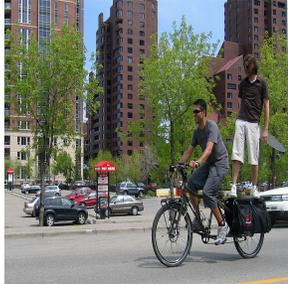

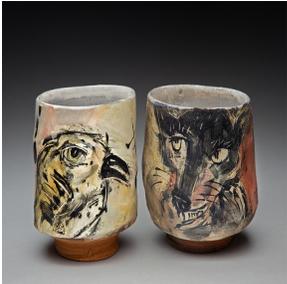 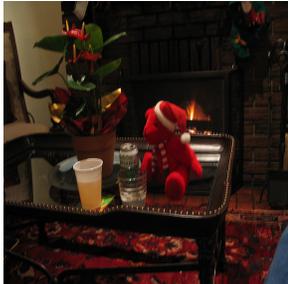 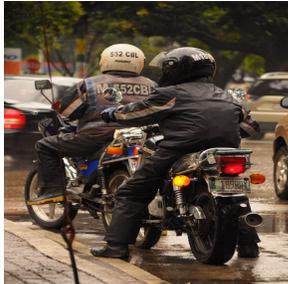 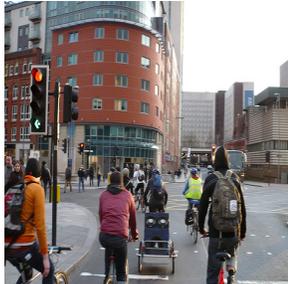

## B  Impact of Perturbations on Image Retrieval Performance

Table 4 lists samples of perturbation effects, with both increases and decreases in R@1.